\pdfoutput=1

\documentclass[11pt]{article}

\usepackage[]{acl}

\usepackage{times}
\usepackage{latexsym}

\usepackage[T1]{fontenc}

\usepackage[utf8]{inputenc}

\usepackage{microtype}

\usepackage{graphicx}
\usepackage{booktabs}
\usepackage{multirow}
\title{Generating Quizzes to Support Training on Quality Management and Assurance in Space Science and Engineering}

\author{Andres Garcia-Silva, Cristian Berrio,  Jose Manuel Gomez-Perez \\
  Expert.ai / Madrid, Spain \\
  \texttt{agarcia@expert.ai}, \texttt{cberrio@expert.ai}, \texttt{jmgomez@expert.ai} \AND
  Jose Antonio Martinez-Heras, Patrick Fleith \\
  Solenix / Darmstadt, Germany \\
  \texttt{jose.martinez@solenix.ch} \And 
  Stefano Scaglioni \\
  ESA / Darmstadt, Germany \\
 \texttt{stefano.scaglioni@esa.int}
  }

\begin{document}
\maketitle
\begin{abstract}
Quality management and assurance is key for space agencies to guarantee the success of space missions, which are high-risk and extremely costly. In this paper, we present a system to generate quizzes, a common resource to evaluate the effectiveness of training sessions, from documents about quality assurance procedures in the Space domain. Our system leverages state of the art auto-regressive models like T5 and BART to generate questions, and a RoBERTa model to extract answers for such questions, thus verifying their suitability. 
\end{abstract}

\section{Introduction}
The complexity, cost, and risk of space missions involving public or private investment and even human lives make quality management a critical requirement to guarantee their success. 
The European Space Agency (ESA) makes a continuous effort to train their staff in quality procedures and standards. Trainees are evaluated to determine the effectiveness of the training sessions, with quizzes as one of the main tools used in such evaluations. 

We present SpaceQQuiz (Space Quality Quiz), a system designed to help trainers to generate quizzes from documents describing quality procedures. Such documents cover topics like \textit{Anomaly and Problem Identification, Reporting and Resolution} or \textit{Configuration Management}, and include stakeholder responsibilities, activities, performance indicators and outputs, among others.   

To design SpaceQQuiz we use state-of-the-art models based on transformers for Question Generation (QG) and Question Answering (QA). Since we could not find specialized models for the space or quality management domains, we reuse models already pre-trained on general-purpose document corpora and fine-tuned on SQuAD\footnote{The Stanford Question Answering Dataset \url{https://rajpurkar.github.io/SQuAD-explorer/}}. 

\begin{figure}[t]
    \centering
    \includegraphics[width=0.5\textwidth]{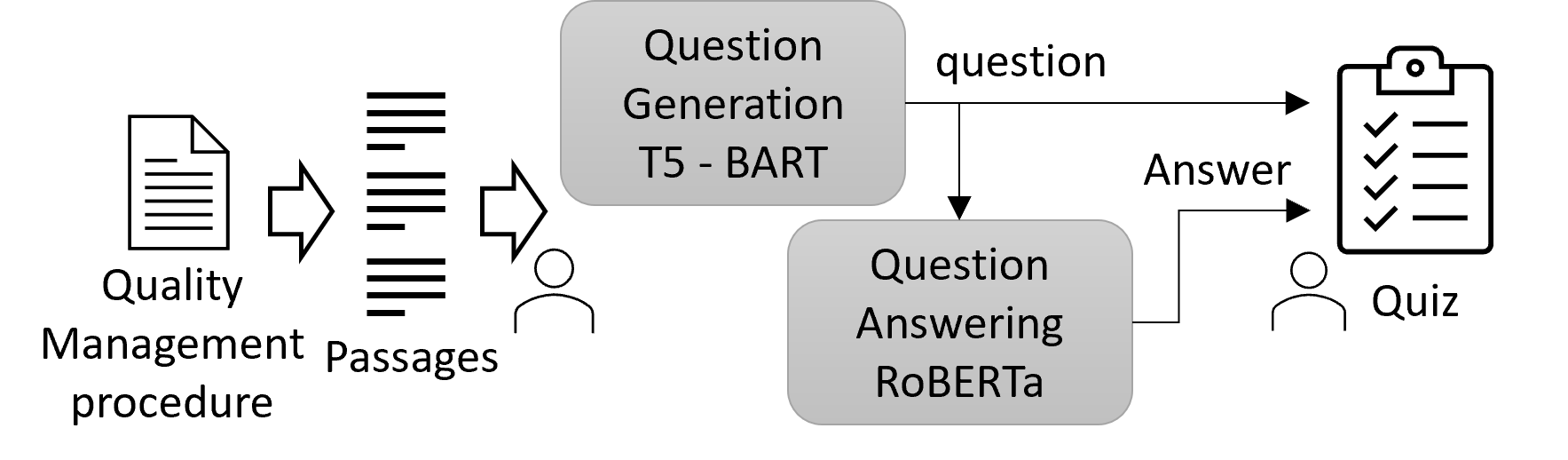}
    \caption{SpaceQQuizz - Proposed architecture.}
   \label{fig:arch}
\end{figure}

\section{Quiz generation system}
Figure \ref{fig:arch} shows the high-level architecture of SpaceQQuiz\footnote{Demo: \url{https://esatde.expertcustomers.ai/SpaceQQuiz/} user/pass demoINLG/demoINLG2022!}. A question generation model is run on each passage extracted from the document. The generated questions and the corresponding passages are fed to a question answering model that extracts the answer from the passage. Only questions with answers are included in the candidate list that then is refined by the trainer to generate the quiz.

The process starts when the trainer uploads a quality procedure document. The system extracts the text from the PDF document using Apache PDFBox\footnote{Apache PDFBox \url{https://pdfbox.apache.org}} and uses regular expressions to identify  sections, subsections and paragraphs while removing non relevant text such as headers and footers
. The trainer is presented with a list of candidate sections so that she can choose the most interesting ones for the quiz. 

\subsection{Question generation}
To generate the questions we use a T5 model \cite{raffelT52020}  and a BART model \cite{lewis-etal-2020-bart} fine-tuned on question generation. We use two models\footnote{Models withdrawn from HuggingFace by their authors. Description available at \url{https://github.com/patil-suraj/question_generation}} in order to increase the number and variety of questions for each text passage. Both T5 and BART have excelled in sequence generation tasks, such as abstractive summarization and abstractive question answering. The models we reuse  were fine-tuned using SQuAD1.1, 
which consists of 100,000 questions created from Wikipedia articles where answers are segments in text passages. 

T5 is fine-tuned using an answer-aware approach where the model is  presented with the answer and a passage to generate the question. T5 is trained on a multitask objective to i) extract answers, ii) generate questions for answers using passages as context, and iii) extract answers for the generated questions. Finally the answer for the generated question is compared with the answer used to generate the questions. BART is fine-tuned following an answer-agnostic approach where the model is trained to generate questions from passages without information about the answers. 

During generation, we use beam search as decoding method, with 5 as number of beams. Beam search keeps the most likely sequence of words at each time step and chooses the final sequence that has the overall highest probability. To avoid duplicity of questions in the final list, we compare them using cosine similarity between the question encoding generated through sentence transformers.\footnote{\url{https://www.sbert.net/}} 
We discard questions similar to a previous one above an empirically defined threshold set at 0.8. 

\subsection{Question answering}
Once the questions have been generated we use a RoBERTa model \cite{liu2019roberta}, fine-tuned for question answering in SQuAD2.0 
to extract answers from the passages. SQuAD2.0 adds 50,000 unanswerable questions to SQuAD1.1. Thus, the fine-tuned RoBERTa is able to generate answers or not depending on the question. If RoBERTa fails to generate an answer for a generated question we remove it from the candidate list of questions presented to the trainer.

\subsection{Quiz generation}
The system displays the list of generated questions, answers, and the passages. The trainer can select specific questions to  include in the quiz. Finally the system generates the quiz with a section containing only the questions to be handed to the trainee and another section reserved for the trainer with questions, answers and passages. 

Table \ref{tab:questions} shows some example questions generated from quality procedure documents. However, a human agent is still necessary since some of the questions generated by the models are too general (e.g. Who is the owner of the system?), syntactically incorrect, or not completely related to the context. More insights are presented in the following section.

\begin{table}[]
\caption{Questions generated by SpaceQQuiz}
\resizebox{0.9\columnwidth}{!}{%
\begin{tabular}{p{\columnwidth}}
\toprule
What is the first source for raising a spacecraft Anomaly Report?               \\
\hspace{3mm} the spacecraft log is the first source for raising ... \\
What does the ARB have to do in case of  an anomaly detected in a shared infrastructure?                                                        \\
\hspace{3mm} notify the relevant infrastructure team                                         \\
Who can issue a supplier waiver?                                                \\
\hspace{3mm} OPS Project   Manager or Service Manager                                          \\
What does the leader of the operator's   team do with the raised Anomaly Reports? \\
\hspace{3mm} performs a preliminary review                                                     \\
Who chairs the   Software Review Board?                                           \\
\hspace{3mm} the owner of   the software,                                                      \\
What is   mandatory for the closure of a Problem Report?                          \\
\hspace{3mm} Root cause   identification                                                       \\
What are minor   non-conformances?                                                \\
\hspace{3mm} by definition, cannot be classified as major.                                   \\
\bottomrule
\end{tabular}%
}
\label{tab:questions}
\end{table}

\section{Evaluation}
To evaluate the quiz generation system, we generate a quiz with 50 questions and answers pairs out of a quality procedure titled \textit{OPS Procedure for Configuration Management}. Then a quality management expert evaluates the generated questions using relevance and correctness as evaluation criteria. The  result of this manual evaluation is reported in table \ref{tab:evaluation}.

\begin{table}[htbp]
  \centering
    \caption{Expert evaluation of question generation and question answering modules. (*) Only answers with a valid question according to the quality expert are evaluated. }
    \begin{tabular}{lr}
    \toprule
          & \multicolumn{1}{l}{Accuracy} \\
    \midrule
    Generated questions & 0.660 \\
    Extracted Answers & 0.600 \\
    Extracted Answers* & 0.818 \\
    \bottomrule
    \end{tabular}%
  \label{tab:evaluation}%
\end{table}%

\begin{table*}[t]
  \centering
  \caption{Example questions rated as non correct in the evaluation. In bold the answers extracted by the question answering module.}
  \resizebox{0.9\textwidth}{!}{%
    \begin{tabular}{cp{0.1\textwidth}p{0.8\textwidth}}
    \toprule
    \multirow{2}{*}{1} & Context & In the process of \textbf{configuration identification} the team shall be aware on what is needed to be put under configuration control. \\
    & Question & What shall the team know on what is needed to be put under configuration control? \\
    \midrule
    \multirow{2}{*}{2} & Context & In a \textbf{continuous service} there is the concept of living baseline over a dynamic scope. \\
    & Question & What is the concept of living baseline over a dynamic scope? \\
    \midrule
    \multirow{2}{*}{3} & Context & Item configuration, in terms of implemented functions (e.g. \textbf{software version 2.0)} \\
    & Question & What is item configuration in terms of implemented functions? \\
    \midrule
    \multirow{2}{*}{4} & Context & The system under configuration includes also \textbf{the items received as Customer Furnished Item.} \\
    & Question & What does the system under configuration include? \\
    \bottomrule
    \end{tabular}%
  }
  \label{tab:wrongquestions}%
\end{table*}%

In total 66\% of questions are considered relevant and grammatically correct, and 60\% of answers are also regarded as correct by the evaluator. If we focus only on the answers of relevant and correct questions then the percentage of accurate answers rises to 81.8\%. The level of accuracy for the question generation makes the human-in-the-loop necessary to guarantee the quality of the questions in the quiz. However, note that the domain expert is also necessary to select the subset of questions to be included in the quiz. 

By analysing incorrect question and answer pairs, we realize that despite being grammatically correct and relevant, some questions are just not possible to answer from the context used to generate them (see for example questions 1 and 2 in the table \ref{tab:wrongquestions}). This is consequence of a failure in the question answering module that produces an answer for such questions. Another example of wrong functioning of the question answering module is shown in question 3 in table \ref{tab:wrongquestions}, where the answer for the given question is extracted from the example in round brackets. A possible solution for this case is to discard  examples in the text before feeding the question generation and the question answering modules. 

For some correct and relevant questions (see for example  question 4 in table \ref{tab:wrongquestions}) the question answering module just return partial answers. In this case the word \textit{also} means that the answer in this context complements the answer already provided in another text excerpt. This a limitation of the extractive question answering module since it only extracts consecutive sequence of tokens from text passages as answers.

Finally the evaluator as domain expert reported that in some cases the problem is the underlying text used to generate the question that is not clear enough to formulate appropriate questions. Thus, wrong questions might indicate text excerpts that need to be reviewed by the author to convey the messages clearly.


\section{Conclusions}
We describe SpaceQQuiz, a system to help quality management trainers in the space domain to speed up the generation of evaluation material. SpaceQQuiz uses autoregresive models such as T5 and BART to generate the questions, and RoBERTa autoencoder to extract answers that are used as indicators of the validity of the questions. 

\section*{Acknowledgements}
This work is funded by ESA under contract AO/1-10291/20/D/AH - “Text and Data Mining to Support Design, Testing and Operations”.

\bibliography{spaceqquizinlg}
\bibliographystyle{acl_natbib}

\end{document}